\documentclass[letterpaper, 10 pt, conference]{ieeeconf}  

\IEEEoverridecommandlockouts                              

\usepackage{graphics} 
\usepackage{epsfig} 
\usepackage{mathptmx} 
\usepackage{times} 
\usepackage{amsmath} 
\usepackage{amssymb}  
\usepackage{algorithm}
\usepackage{algpseudocode}
\usepackage{bm}
\usepackage{booktabs}
\let\labelindent\relax
\usepackage{enumitem}
\usepackage{placeins}
\usepackage{subfig}
\usepackage{hyperref}

\title{\LARGE \bf
Physics-informed Imitative Reinforcement Learning for Real-world Driving
}

\author{Hang Zhou$^{1}$, Yihao Qin$^{1}$, Dan Xu$^{2}$ and Yiding Ji$^{1}$
\thanks{$^{1}$Robotics and Autonomous Systems Thrust, Systems Hub,
            The Hong Kong University of Science and Technology (Guangzhou), Guangzhou, China
        {\tt\small hzhou269@connect.hkust-gz.edu.cn}; {\tt\small jiyiding@hkust-gz.edu.cn}}%
\thanks{$^{2}$Department of Computer Science and Engineering, School of Engineering, The Hong Kong University of Science and Technology, Hong Kong SAR, China 
        {\tt\small  danxu@cse.ust.hk}}
}

\begin{document}

\maketitle
\thispagestyle{empty}
\pagestyle{empty}

\begin{abstract}
Recent advances in imitative reinforcement learning (IRL) have considerably enhanced the ability of autonomous agents to assimilate expert demonstrations, leading to rapid skill acquisition in a range of demanding tasks. However, such learning-based agents face significant challenges when transferring knowledge to highly dynamic closed-loop environments. Their performance is significantly impacted by the conflicting optimization objectives of imitation learning (IL) and reinforcement learning (RL), sample inefficiency, and the complexity of uncovering the hidden world model and physics. To address this challenge, we propose a physics-informed IRL that is entirely data-driven. It leverages both expert demonstration data and exploratory data with a joint optimization objective, allowing the underlying physical principles of vehicle dynamics to emerge naturally from the training process. The performance is evaluated through empirical experiments and results exceed popular IL, RL and IRL algorithms in closed-loop settings on Waymax benchmark. Our approach exhibits 37.8\% reduction in collision rate and 22.2\% reduction in off-road rate compared to the baseline method. 


\end{abstract}

\section{INTRODUCTION}
\vspace{-3pt}
\label{sec:intro}
Imitative reinforcement learning (IRL) combines demonstration-based and reward-driven objectives and have significantly improved upon both imitation learning (IL) and reinforcement learning (RL). Such methods offer enhanced stability, implicit reward modeling, and context-aware decision making. Early approaches, such as behavior cloning followed by RL fine-tuning provide a sample-efficient initialization by directly mimicking expert behavior, while later reinforcement phases refine these behaviors toward improved performance. Subsequent methods advanced imitative reinforcement learning by directly embedding expert guidance into the RL objective. DQfD \cite{DQfD} improves sample efficiency and accelerates early learning by integrating supervised imitation losses within a value-based RL framework. DAPG \cite{Rajeswaran-RSS-18} utilizes a two-stage approach: it initially pre-trains the policy using behavior cloning to establish an expert-like baseline, then performs RL fine-tuning augmented with an imitation learning loss to ensure continued adherence to expert behaviors. In contrast, BC-SAC \cite{10342038} unifies the imitation and reinforcement learning objectives into a single-stage loss function, thereby streamlining the learning process without distinct pre-training and fine-tuning phases.


\begin{figure}[t]
\centering
\includegraphics[ width=1.0\linewidth]{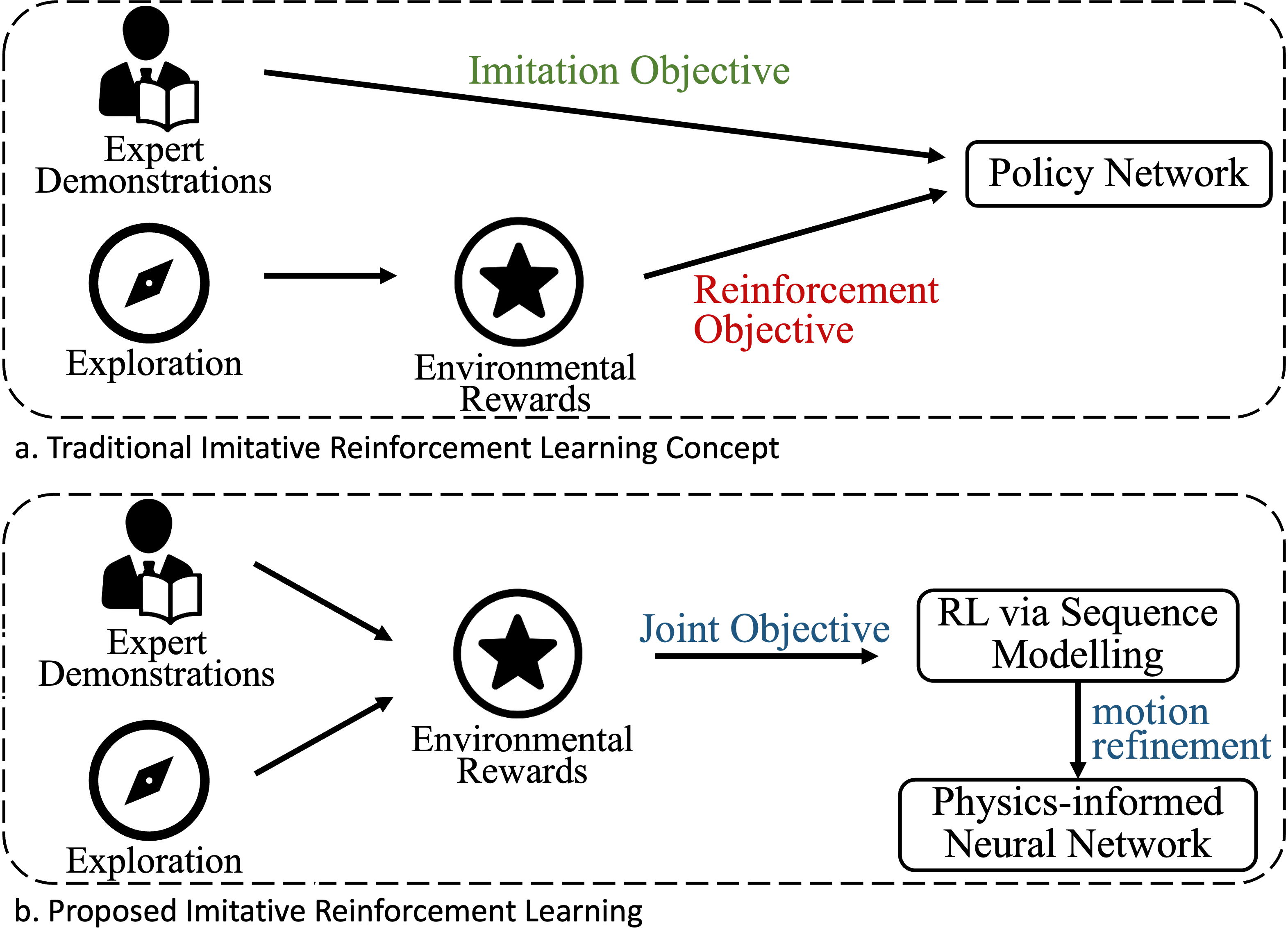}
\vspace{-15pt}
\caption{The upper figure depicts traditional IRL, highlighting its limitations in conflicting optimization goal with imitation objective and reinforcement objective. The bottom figure presents our IRL pipeline that jointly optimizes with sequenced modelling based reinforcement learning approach, and later refined by physics-informed neural network.}
\label{fig:irl}
\vspace{-21pt}
\end{figure}

However, A primary concern of IRL is the misalignment of objectives between the IL and RL phases \cite{leiva2024combining} as described in Fig.~\ref{fig:irl}(a). IL focuses on replicating expert behavior, whereas RL seeks to maximize long-term rewards through exploration. This discrepancy can result in compounding errors especially when the policy encounters unfamiliar states during RL fine-tuning \cite{ross2011reduction,cheng2018bridging}. Furthermore, there is a risk of catastrophic forgetting, where essential expert behaviors learned in the IL phase may be lost by the emphasis on reward maximization \cite{kaplanis2018catastrophic} during the RL phase. In highly varying and dynamic autonomous driving scenarios, the distribution shifts problem needs more attention as generalizing an RL agent from one scenario can lead to divergence in another. Previous work \cite{10342038, DQfD} explored the joint optimization of IL and RL through a unified loss function. However, balancing the conflicting optimization goals of IL and RL remains a significant challenge \cite{leiva2024combining}. This is because IL typically relies on supervised loss functions which aim to mimic expert behavior, while RL employs reward-driven learning and focuses on maximizing cumulative rewards.


The deployment of imitative reinforcement learning in autonomous driving accentuates both algorithmic and safety challenges. RL often faces significant challenges due to a limited understanding of physics, particularly in applications like autonomous driving \cite{8951131,xu2021moving}. These systems rely on expert demonstrations to guide behavior, yet they lack the ability to accurately interpret and respond to complex physical scenarios such as near-collisions, severe weather conditions, and atypical traffic patterns. This gap in understanding physical dynamics and scenarios can hinder the adaptability of the system in real-world environments, leading to increased safety risks. 


 This paper seeks to address the challenges by introducing Physics-informed Decision Transformer (PiDT) and online imitative reinforcement learning (OIRL) pipeline. PiDT effectively transforms large-scale demonstration learning problem into RL tasks through sequence modeling with proposed OIRL pipeline. During the online adaptation stage, network jointly optimizes both the expert trajectory and the exploration trajectory without introducing conflicting optimization goals between IL and RL. Physics-informed network is designed to monitor the energy flow of the ego-vehicle and provide action refinement for the output of sequence modeling. Experiment results indicate that our approach yields substantial performance enhancement in terms of policy performance and sample efficiency in closed-loop settings. PiDT exhibits 37.8\% reduction in collision rate and 22.2\% improvement in reaching the destination compared to the baseline method. This work is conducted entirely with Jax, which facilitates highly efficient training in large-scale data and real-time inference. The inference time is 1.65 milliseconds for PiDT (median) on RTX 3090. The main contributions of this work are summarized as follows:
\begin{itemize}[leftmargin=*]
\item We present a novel imitative reinforcement learning architecture that seamlessly injects expert driving prior into RL without introducing conflicting optimization objectives. 
\item We design a port-hamiltonian neural network to model kinetic energy flow and refine the Decision Transformer’s motion output, thereby improving long-horizon motion prediction and learning of physics information.
\item We introduce a hybrid experience sampling strategy to sequence modeling reinforcement learning and enable sample-efficient training for large-scale datasets.
\end{itemize}

\section{RELATED WORK}
\vspace{-3pt}
\label{sec:relatedwork}

\noindent\textbf {Reinforcement Learning via Sequence Modeling}. Trajectory Transformer \cite{janner2021sequence} and Decision Transformer (DT) \cite{NEURIPS2021_7f489f64} are pioneers in this area, leveraging transformer architectures to model sequences of state-action-reward trajectories and predicting future actions in an offline manner. The following work \cite{lee2022multigame, meng2022offline, wu2023elastic, NEURIPS2023_f58c2479} extends the power of transformers to efficient and generalized decision making in RL. Online DT \cite{zheng2022online} and Hyper DT \cite{xu2023hyperdecision} adapt the original concept to online settings and interact with environments. However, previous work are done on relatively simple environments compared to autonomous driving environments.

\noindent\textbf {Learning with Real-world Driving Data.}
Much work has been done to accommodate with real-world driving data \cite{Sun_2020_CVPR,ricci2018monocular,nuplan,lyft2020,mei2025hamf,10160449} for generalizable driving policy. BC-SAC \cite{10342038} explores the cooperation between reinforcement learning and imitation learning in terms of loss design for real-world driving data. TuPlan \cite{Dauner2023CORL} combines both learning methods with rule-based methods for real-world planning. Guided Online Distillation \cite{li2023guidedonlinedistillationpromoting} adapts real-world data for reinforcement learning online distillation in MetaDrive Simulator~\cite{li2022metadrive}. Trajeglish~\cite{philion2024trajeglish} models real-world traffic autoregressively as a language processing problem with a causal transformer. Our approach is most similar to Trajeglish as both works tokenize expert driving logs to state-action sequences, encode agent and map information for better scene understanding and finally output actions for agent control. However, Trajeglish is fundamentally a supervised learning method and it inevitably has training and test distribution mismatch. Our approach differs as PiDT additionally models return to state-action sequences and convert supervised learning to RL via sequence modeling.

\noindent\textbf {Hamiltonian Neural Networks (HNNs)}~\cite{greydanus2019hamiltonian} capture the dynamics of conservative systems by learning an underlying Hamiltonian. However, many real-world cases involve energy flow, injection, and loss. To accommodate such non-conservative dynamics, Port-Hamiltonian Neural Networks~\cite{desai2021port} augment the HNN framework by explicitly modeling energy ports, while Dissipative Hamiltonian Neural Networks \cite{sosanya2022dissipative} incorporate dissipation terms to reflect friction and drag. Together, these approaches extend traditional modeling methods, providing robust tools for accurately simulating systems with energy exchange. 
\vspace{-3pt}

\section{PRELIMINARY}
\vspace{-3pt}
\noindent\textbf {Port-Hamiltonian Neural Networks (PHNN)}. 
In many practical systems such as vehicles, external actuators allow energy to flow into and out of the system. Classical Hamiltonian mechanics assumes energy conservation, where the position variable ($q$) and momentum variable ($p$) evolve as
\begin{equation}
\dot{q} = -\frac{\partial H}{\partial p}, \ \ \ \ \dot{p} = -\frac{\partial H}{\partial q}
\label{eq:classical-momentum}
\end{equation}
with \(H\) denoting the Hamiltonian (total energy). To model nonconservative phenomena (e.g., engine thrust and brake), the momentum equation is augmented with a generalized force $
Q_{\text{nc}} = F_{\text{extern}} - F_{\text{drag}}$,
with \(F_{\text{extern}}\) representing the injected force and \(F_{\text{drag}}\) capturing energy losses.
Therefore, the momentum equation turns to $\dot{p} = -\frac{\partial H}{\partial q} + Q_{\text{nc}}$.

A unified description of energy exchanges is provided by the port-Hamiltonian framework. Defining the state vector as
$x = \begin{pmatrix} q \\ p \end{pmatrix}$,
the system dynamics are given by
\begin{equation}
\dot{x} = \Bigl[J(x) - R(x)\Bigr] \nabla H(x) + G(x) u
\end{equation}
\(J(x)\) is a skew-symmetric interconnection matrix, \(R(x)\) is a symmetric positive semidefinite dissipation matrix, \(G(x)\) is an input mapping matrix, and \(u\) denotes external inputs (such as the engine or brake force). Here the conservative potential is neglected (\(\partial H/\partial q \approx 0\)), the dynamics reduce to
$\dot{q} = \frac{p}{m}, \dot{p} = F_{\text{extern}} - F_{\text{drag}} \label{eq:veh-p}$, with the Hamiltonian simplified to $H = \frac{p^2}{2m}$, which represents system's kinetic energy.

\section{METHOD}
\begin{figure*}[t]
\centering
\includegraphics[ width=1\linewidth]{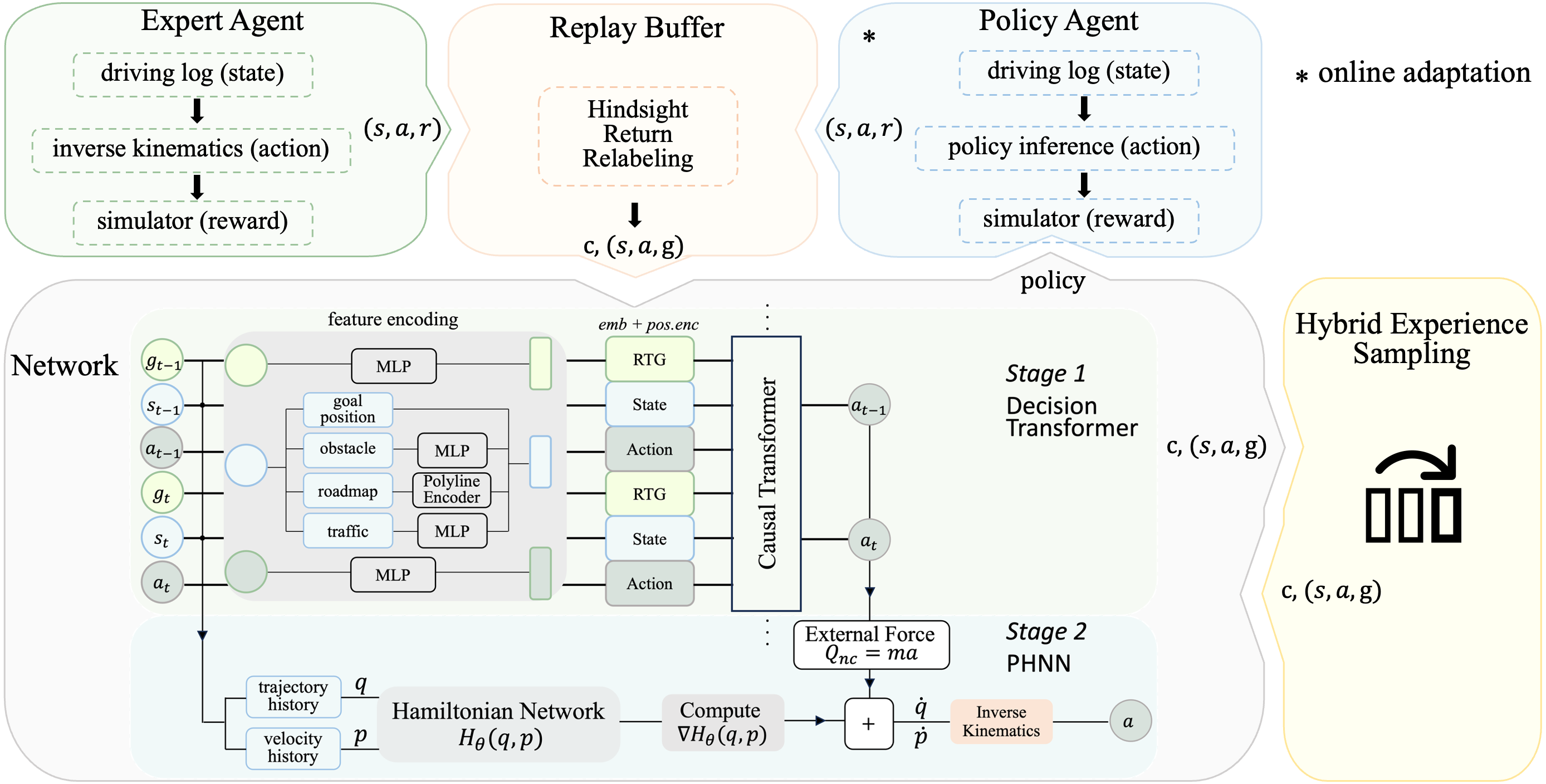}
\vspace{-16pt}
\caption{The general training procedure of PiDT. Real-world driving scenarios are reproduced in the simulator, the produced transitions $(s, a)$ and assigned reward are recorded into replay buffer. $HindsightReturnRelabeling$ is performed to transform transitions to state-action-return pairs $(s, a, g)$ for sequence modelling. During training, hybrid experience sampling is performed for sample efficiency. In the first stage of training, the Decision Transformer (DT) is trained solely on expert demonstrations. The second stage of training have additional online adaptation data and the output of DT to jointly train the entire physics-informed network (DT+PHNN).} 
\label{fig:network_architecture}
\vspace{-18pt}
\end{figure*}

In this section, we present PiDT, a novel two-stage imitative reinforcement learning framework designed for dynamic driving scenarios. PiDT adapts a Decision Transformer for nuanced decision-making, an online imitative reinforcement learning pipeline for continuous adaptation and improvement, a hybrid experience sampling mechanism to enhance learning efficiency, and second stage port-hamiltonian neural network for physics-informed motion refinement. 
\vspace{-3pt}

\subsection{Network Design for PiDT}
\vspace{-3pt}
PiDT utilizes a causal transformer for sequence modeling of the traffic, vehicle control action and corresponding $return\_to\_go$. The feature encoder is specifically designed for autonomous driving scenarios. 
Given that the real-world driving environment is complex and dynamic, a dedicated feature encoding network is developed for state representation. The real-world driving state contains perceptual information such as obstacles, road map, traffic and so on. We follow a vectorized representation to organize the road map as polylines, which are then extracted using Polyline Encoder \cite{shi2022motion}. Obstacles with past 10 historical information are recorded in terms of $[position\_x,\ position\_y,\ yaw,\ speed,\ length,\ width]$. Obstacle and traffic embedding are extracted with multi-layer perception networks. 

The work is extended to goal-conditioned reinforcement learning by adding the relative vector goal between the ego vehicle and the destination in terms of ($x, y$). The importance of goal condition lies in its influence on the decision-making process of the autonomous agent. Even in an identical environment, the actions taken by the vehicle can vary significantly depending on the specified goal position. 

\noindent\textbf{Physics Informed Network for Decision Transformer.} 
Port-Hamiltonian Neural Network is applied for motion refinement in trajectory level. We design the PHNN to not only reconstruct the underlying kinetic energy landscape but also account for energy injections due to external forces, thereby providing a robust framework for systems like vehicles where nonconservative dynamics are prominent. The final learned position output is converted to action by inverse kinematics.


In modern automotive applications, particularly within autonomous driving systems, advanced control techniques maintain vehicles at near-constant speeds by mitigating the impact of dissipative forces such as road friction and aerodynamic drag. This focus shifts attention to the energy exchanges associated with variations in kinetic energy, where the dominant dynamic contributions stem from engine thrust and braking, manifesting as changes in acceleration. Consequently, the external forcing term \( Q_{\text{nc}} \) is modeled via \( F = m\cdot a \). Within the Port-Hamiltonian Neural Network framework, the Hamiltonian is predominately driven by the kinetic energy component, while external forcing is leveraged to capture the acceleration effects. By minimizing the influence of lower-level dissipative forces, the model concentrates on the primary energy exchanges due to engine and brake actions, thereby providing a robust description of the vehicle's dynamic behavior.

The objective is to have a PHNN that learns both the intrinsic energy dynamics and the nonconservative effects from data. The proposed approach comprises two primary components:
(1) A neural approximation \(H_\theta(q,p)\) of the conservative energy function.
(2) A learned correction term \(Q_{\text{nc}}= m\cdot a\) that captures the nonconservative forces. Here the action $(a)$ is the output of pretrained Decision Transformer. The learned dynamics are modeled accordingly as:
\begin{align}
\dot{q} = \frac{\partial H_\theta(q,p)}{\partial p},  \ \ \ \ \
\dot{p} = -\frac{\partial H_\theta(q,p)}{\partial q} + Q_{\text{nc}}
\label{eq:phnn-p}
\end{align}
The entire system may be encapsulated in the unified form:
\begin{equation}
\dot{x} = \underbrace{J\nabla H_{\theta}(q,p)}_{\text{Conservative Structure}} + \underbrace{Q_{nc}}_{\text{External Forcing}}
\label{eq:phnn-full}
\end{equation}
where \(J\) is a fixed skew-symmetric matrix that encodes the system's conservative dynamics. This formulation ensures that both the underlying conservative structure and the external forcing are simultaneously and effectively modeled.


\vspace{-3pt}

\subsection{Online Imitative Reinforcement Learning}
\vspace{-3pt}
The proposed online imitative reinforcement learning pipeline (Alg.~\ref{oirl}) begins by rolling out expert driving log data (state) while applying inverse kinematics (action) and reward shaping (reward) for off-policy pre-training. This initial phase is designed to quickly shift the distribution toward expert behavior. At the midpoint of training, the model undergoes a mixed on-policy adaptation (OPA) phase to reduce environmental distribution shifts. Notably, the network is trained solely on the (state, action, return\_to\_go (g)) pairs without incorporating a separate RL optimization objective. In this framework, the network optimizes to execute actions associated with high g values while avoiding those linked with low g values, and both OPA and offline expert rollouts are conducted concurrently post mid-training to prevent convergence to local minima.

The primary objective of imitative reinforcement learning is to fully leverage an extensive expert dataset to enhance policy robustness within realistic simulation environments. This is achieved by adopting principles akin to shaped imitation learning (SIL) \cite{Judah_Fern_Tadepalli_Goetschalckx_2014} and GRI \cite{chen2021gri}, where the reward structure is tailored based on expert demonstration data. State-action pairs that adhere closely to expert trajectories receive a positive reward and result in higher return-to-go values. Unlike Behavior Cloning \cite{argall2009survey}, which primarily learns state-action mappings, the network integrates state-action-return triplets, effectively shifting its learning paradigm into reinforcement learning. Expert driving trajectories are converted to actions via inverse kinematics, with negative rewards penalizing off-road and overlap (collision) behaviors. 

Initially, two replay buffers are set up: a Transition Replay Buffer ($D\_{trans}$) with a specified capacity (A), and a Trajectory Replay Buffer ($D\_{traj}$). The algorithm operates over a number of scenarios ($num\_scenarios$). In the first phase, called Online Data Collection, the algorithm populates ($D\_{trans}$) with state-action-reward tuples (($s, a, r$)). For the first half of the scenarios, it relies solely on human expert driving data to fill the buffer. In the latter half, it combines this expert data with exploratory data generated by the policy agent ($\pi_\theta$). Once the buffer is full, the algorithm performs a process called Hindsight Return Relabeling to transfer and transform data from transition level ($D\_{trans}$) to trajectory level ($D\_{traj}$). The algorithm then enters a training loop, iterating 1000 times to sample and shuffle obstacle orders from ($D\_{traj}$), which provides varied and rich experiences for training the policy agent. Moreover, the dataloader randomly shuffles ($ShuffleObstacleOrder$) the obstacles within each state to boost data augmentation.

\begin{algorithm}[!t]
\caption{Online Imitative Reinforcement Learning}\label{pipeline}
\begin{algorithmic}
\State Initialize Transition Replay Buffer $D_{trans}$ for capacity A, Trajectory Replay Buffer $D_{traj}$ 
\While{$n \le num\_scenarios$}
\While{$D_{trans}$ is not full}        \Comment{Online Data Collection}
\If{$n \le 0.5*num\_scenarios$}
    \State reproduce scenarios with Human Expert Driving Data,  $D_{trans} \gets (s,a,r)$
\Else
    \State reproduce scenarios with Human Expert Driving Data, $D_{trans} \gets (s,a,r)$
    \State explore scenarios with Policy agent $\pi_\theta$ , $D_{trans} \gets (s,a,r)$
\EndIf
\EndWhile
\State \text{$HindsightReturnRelabeling$:} $D_{traj}\gets D_{trans}$ $\equiv$ $ [[(s_{i,j}, a_{i,j}, g_{i,j})]_{i=0}^{T}]_{j=0}^{A/T} \gets  \left[(s_i, a_i, r_i) \right]_{i=0}^{A}$
\State $D_{trans} \gets \emptyset$
\For {k in range(1000)}: 
\State sample and $ShuffleObstacleOrder$: $ [[(s_{i,j}, a_{i,j}, g_{i,j})]_{i=t-c}^{t}]_{j=0}^{B} \gets D_{traj}$
\State train on sampled data
\EndFor
\EndWhile
\end{algorithmic}
\label{oirl}
\end{algorithm}

\begin{table*}[t]
\footnotesize
\setlength{\tabcolsep}{1mm}
\centering
\begin{tabular}{p{3.0cm} p{1.5cm}p{1.15cm}|p{1.7cm}p{1.7cm}p{1.9cm}p{1.5cm}p{1.8cm}}
\specialrule{.15em}{.1em}{.1em}
Agent & Action Space & Train Sim Agent & Off-Road Rate (\%) $\downarrow$ & Collision Rate (\%) $\downarrow$ & Kinematic Infeasibility(\%)$\downarrow$ & ADE (m) & Route Progress Ratio (\%) \\
\hline

Wayformer \cite{nayakanti2022wayformer} & Delta  & -  & 7.89 & 10.68 & 5.40 & \textbf{2.38} & 123.58 \\

LatentDriver \cite{xiao2025latentdriver}  &  Delta  & -  & \textbf{2.33$\pm$0.13} & 3.17$\pm$0.04 & 4.78 & - &\textbf{99.57$\pm$0.1} \\

BC \cite{argall2009survey} & Delta & -& 4.14$\pm$2.04 & 5.83$\pm$1.09 & 0.18$\pm$0.16 & 6.28$\pm$1.93 & 79.58$\pm$24.98 \\ 

BC \cite{argall2009survey} & Delta (D) & -& 4.42$\pm$0.19 & 5.97$\pm$0.10 & 66.25$\pm$0.22 & 2.98$\pm$0.06 & 98.82$\pm$3.46 \\ 

BC \cite{argall2009survey} & Bicycle & -& 13.59$\pm$12.71 & 11.20$\pm$5.34 & 0.00$\pm$0.00 & 3.60$\pm$1.11 & 137.11$\pm$33.78 \\



\hline
DQN \cite{mnih2013playing} & Bicycle (D) & IDM & 3.74$\pm$0.90 & 6.50$\pm$0.31 & 0.00$\pm$0.00 & 9.83$\pm$0.48 & 177.91$\pm$5.67 \\ 
DQN  \cite{mnih2013playing} & Bicycle (D) & Playback & 4.31$\pm$1.09 & 4.91$\pm$0.70 & 0.00$\pm$0.00 & 10.74$\pm$0.53 & 215.26$\pm$38.20 \\
EasyChauffeur-PPO \cite{xiao2024easychauffeur} & Bicycle & Playback & 3.95 & 4.72 & 0.00 & - & 98.26\\
DT \cite{NEURIPS2021_7f489f64} & Bicycle & Playback &   6.21$\pm$0.35 &  3.62$\pm$0.18 &  0.00$\pm$0.00 &  8.32$\pm$0.68 & 116.78$\pm$7.43 \\
OnlineDT \cite{zheng2022online}& Bicycle & Playback &  3.92$\pm$0.25 &  3.05$\pm$0.14 &  0.00$\pm$0.00 & 7.78$\pm$0.60 & 95.96$\pm$9.29 \\
DQfD \cite{DQfD} & Bicycle (D) & Playback & 3.83$\pm$0.96 & 2.99$\pm$0.81 & 0.00$\pm$0.00 & 5.3$\pm$1.35 & 98.77$\pm$12.53\\
\textbf{PiDT (ours)} & Bicycle & Playback &  2.98$\pm$0.23    &  \textbf{1.86$\pm$0.08}      & \textbf{0.00$\pm$0.00} &  6.99$\pm$0.69   & 102.41$\pm$4.66 \\

\specialrule{.15em}{.1em}{.1em}
\end{tabular}
\vspace{-5pt}
\caption{Closed-loop Benchmark. Performance evaluations are done against IDM simulation agents. Agents run without any termination conditions. Models report mean and standard deviation over 3 seeds. Action space is continuous unless denoted with D (discrete). Route Progress Ratio exceeds 100\% when the ego vehicle overshoots its target goal position.} 
\label{table:baseline_agents}
\vspace{-10pt}
\end{table*}

\begin{table}[t]
\footnotesize
\setlength{\tabcolsep}{1mm}
\centering
\begin{tabular}{p{2.0cm} p{1.3cm} | p{2.0cm} p{2.0cm}}
\specialrule{.15em}{.05em}{.05em}
Method & Train Sim Agent & Failure Rate (\%) & Route Progress Ratio (\%) \\
\hline
BC \cite{argall2009survey} & - & 4.35$\pm$0.27 & 99.00$\pm$0.39 \\ 
SAC \cite{haarnoja2018soft} & Playback & 6.66$\pm$0.44 & 77.82$\pm$8.21 \\
BC-SAC \cite{10342038} & Playback & 3.35$\pm$0.31 & 95.26$\pm$8.64 \\
\textbf{PiDT (ours)} & Playback & 3.17$\pm$0.22 & 104.63$\pm$2.63 \\
\specialrule{.15em}{.05em}{.05em}
\end{tabular}
\vspace{-5pt}
\caption{Closed-loop Benchmark against playback agent. The failure rate (lower is better) is the percentage of the run segments that have collision or off-road event.} 
\label{table:open_loopresults}
\vspace{-15pt}
\end{table}

\noindent\textbf{Reward Function Definition}:
\begin{equation}\label{RI}
R_{imitaiton} = \begin{cases}
1.0 & \text{if } \text{{log\_divergence}} < 0.2, \\
0.0 & \text{if } \text{{log\_divergence}} > 0.2, 
\end{cases} 
\end{equation}
\begin{equation}\label{RO}
R_{off\_road} = -2
\end{equation}
\begin{equation}\label{RC}
R_{overlap} = -10 
\end{equation}
where {log\_divergence} is the euclidean distance between the expert log history of the ego vehicle and that of the controlled agent. The imitation term \(R_{\text{imitation}}\) in Eq. (\ref{RI}) rewards behaviors that closely align with expert driving, and avoids large rewards to encourage exploration around expert trajectories. The threshold of 0.2m is chosen to reward actions closely following expert demonstrations. However, we consider 0.2m to be the maximum that we can set as values larger than 0.2m can lead to collisions and out-of-lane actions which should not be rewarded. Safety is enforced through the penalties \(R_{\text{off\_road}}=-2\) and \(R_{\text{overlap}}=-10\), where the latter is set five times larger to reflect the higher real-world severity of collisions relative to brief lane excursions. Furthermore, we set the absolute magnitude of \(R_{\text{imitation}}\) much lower than that of \(R_{\text{overlap}}\) to encourage the agent not only to imitate expert behavior but also to learn a general-purpose driving policy. This design choice acknowledges that there is not only one viable trajectory to reach a destination.


\subsection{Hybrid Experience Sampling for Decision Transformer}
\vspace{-3pt}

The proposed imitative reinforcement learning method not only handles large volumes of offline demonstration data but also obtains an infinite amount of data through online interactions. The ability to utilize prioritized experience replay (PER) becomes critical for improving sample efficiency in reinforcement learning. The original proposed PER \cite{schaul2015prioritized} selectively samples experiences with high temporal-difference errors from the replay buffer to focus on more informative experiences. However, DT does not use temporal-difference errors and therefore precludes direct application of PER. Instead, we adapt by using action loss to gauge transition importance within the Decision Transformer, which assesses state-action-return relationships. The design concept is that if the model's predicted actions diverge from actual ones, it indicates a misinterpretation of the environment.

We develop hybrid experience sampling (HES) for Decision Transformer on top of the previous online imitative training pipeline (Alg.~\ref{oirl}). Extra hybrid trajectory replay buffers ($D_{single}^{hes}$, $D_{overall}^{hes}$) are designed to store important sampled trajectories based on action loss. The action loss represents the difference between the actions predicted by the policy network and the actual actions taken. A low actor loss means that the policy network's predictions are close to the actual actions, while a high actor loss means that the predictions are far from the actual actions. Different from bootstrapping RL algorithms which sample one pair of state, action and reward $(s_{i}, a_{i}, r_{i})$, Decision Transformer samples a context length($c$) of state-action-return pairs $[[(s_{i,j}, a_{i,j}, g_{i,j})]_{i=t-c}^{t}]_{j}$ for sequence modeling. Therefore, the replay buffers designed for the decision transformer store a set of state-action-return pairs with a specified context length. Our proposed prioritization rule takes advantage of these state-action-return pairs, which try to analyze and understand either single scene or cumulative scenes. The proposed method prioritizes trajectories based on the following criteria:

\noindent\textbf{Criterion 1:} Preservation of transitions which contains single-step action discrepancy ($L_{a}$): The method concentrates on isolating the instances where the model's prognostications manifest the greatest deviation from expected accuracy. Such a strategy is instrumental in directing the model's learning efforts toward ameliorating its most significant errors. The replay buffer for single-step action discrepancy $D_{single}^{hes}$ generally hold unexpected collisions and out-of-line actions.
\begin{equation}\label{hes_s}
D_{single}^{hes} \gets \{ [[(s_{i,j}, a_{i,j}, g_{i,j})]_{i=t-c}^{t}]_j, L_{a} \}
\end{equation}

\noindent\textbf{Criterion 2:}\phantomsection \label{criterion2} Preservation of transitions with maximum cumulative action discrepancy ($L_{ma}$): This method is characterized by its emphasis on identifying and retaining sequences in which the aggregate error of the model's predictions reaches its apex. It holds particular utility for endeavors aimed at refining the model's performance across a continuum of actions for context length of state-action-return pairs. The replay buffer for cumulative action discrepancy $D_{overall}^{hes}$ is useful for preserving long-tail scenarios.
\begin{equation}\label{hes_c}
D_{overall}^{hes} \gets \{ [[(s_{i,j}, a_{i,j}, g_{i,j})]_{i=t-c}^{t}]_j, L_{ma} \}
\end{equation}





The two replay buffers store data based on high value in low value out. The hybrid trajectory replay buffers are sampled for training every fixed amount of episodes and the priorities are updated in the meantime. The goal for the proposed hybrid experience sampling in this paper is to prioritize the trajectories where the model has the biggest misunderstanding of the corresponding scenarios, and therefore to prioritize long-tail scenarios. The final training pipeline is illustrated in Fig.~\ref{fig:network_architecture}.

\section{EXPERIMENTAL RESULTS}
\label{sec:result}

\begin{figure*}
\centering
\includegraphics[width=1.0 \linewidth]{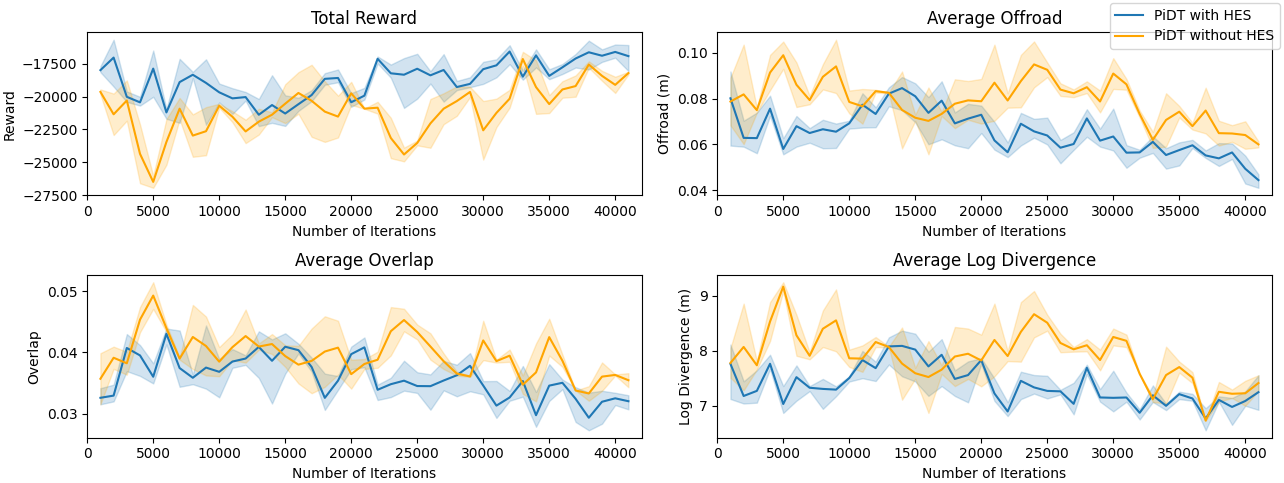}
\vspace{-15pt}
\caption{Comparison of learning curves for PiDT(small) settings with and without Hybrid Experience Sampling. The metric for reward indicates that higher values are better, whereas for off-road incidents, overlap, and log divergence, lower values are better. Models report mean and standard deviation over 3 seeds.}
\label{fig:8}
\vspace{-15pt}
\end{figure*}

\subsection{Experimental Setup}
\vspace{-3pt}
\noindent\textbf{Dataset, Simulator, and Metrics}. Training and experiments are conducted using the Waymo Open Dataset and the Waymax simulator. Waymax provides embedded support for reinforcement learning and diverse scenarios drawn from real driving data. It integrates with the Waymo Open Motion Dataset (WOMD), which offers 531,101 real-world driving scenarios for training and 44,096 scenarios for validation. Each scenario contains 90 frames of data. Specifically, WOMD v1.2 and the exact same metrics (off-road rate, collision rate, kinematic infeasibility, average displacement error (ADE), route progress ratio) from Waymax \cite{gulino2023waymax} are used for benchmarking with the paper. Evaluations are conducted in closed-loop settings against both reactive agents and non-reactvie agents for a wider range of performance comparisons. The training uses 20\% training dataset and evaluations are done on the entire validation dataset.


\noindent\textbf{Implementation Details}. Models of various sizes are developed to perform ablation studies and assess the final performance effectively. Raw observation takes ego vehicle, 15 nearest dynamic obstacles, 300 of closest roadgraph elements, traffic signals and position goal as input. The total size for each step observation is $8892$ and feature extraction is applied to reduce the total size. PiDT(small) has 384 tokens for each element of $(s, a, g)$ pair, 10 blocks, 16 attention heads and in total 22.8 million parameters. PiDT(small) has 512 tokens for each element of $(s, a, g)$ pair, 15 blocks, 16 attention heads and in total 53.4 million parameters. Models use context length with value 10 and enable causal transformer accessing to past 10 $(s, a, g)$ pairs.

\begin{table}[t]
\footnotesize
\setlength{\tabcolsep}{1mm}
\centering
\begin{tabular}{p{3.5 cm} |p{1.35cm}p{1.35cm}p{1.2cm}}
\specialrule{.15em}{.05em}{.05em}
Agent  & Off-Road Rate (\%) & Collision Rate (\%)  & ADE (m) \\
\hline 
PiDT  &    6.21$\pm$0.35    &    3.62$\pm$0.18    &    8.32$\pm$0.68   \\
PiDT + OIRL  &  4.32$\pm$0.27 &  3.21$\pm$0.17 &    8.17$\pm$0.72 \\
PiDT + OIRL  + HES  &  3.52$\pm$0.26   &   2.69$\pm$0.10      &  7.14$\pm$0.63 \\
PiDT + OIRL + HES + PHNN  &  2.98$\pm$0.23   &   1.86$\pm$0.08       &  6.99$\pm$0.69  \\
\specialrule{.15em}{.05em}{.05em}
\end{tabular}
\vspace{-5pt}
\caption{Ablation Study in Closed-loop setting} 
\label{table:ablation}
\vspace{-15pt}
\end{table}

\begin{figure*}[t]
\centering
\subfloat[Irregular Traffic Junction\label{fig:4-1}]{%
\includegraphics[width=0.29\linewidth]{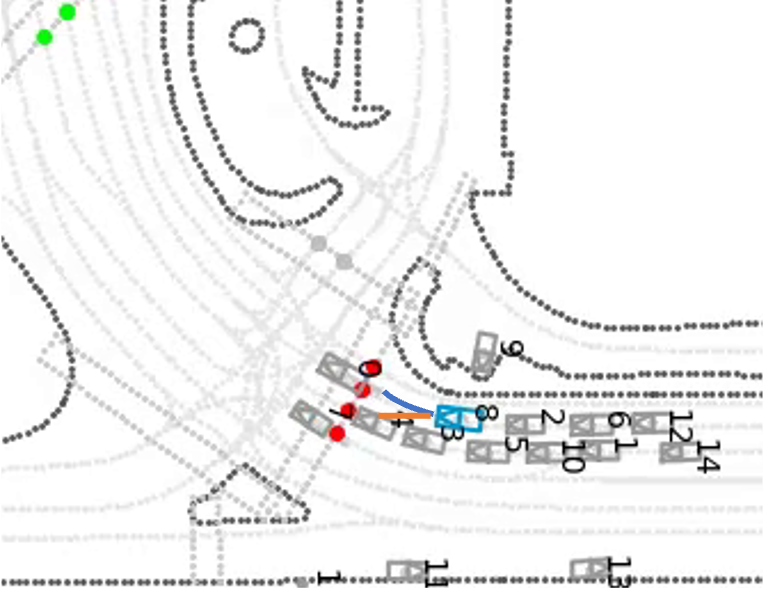}
}
\hfill
\subfloat[Parking Slot\label{fig:4-2}]{%
\includegraphics[width=0.29\linewidth]{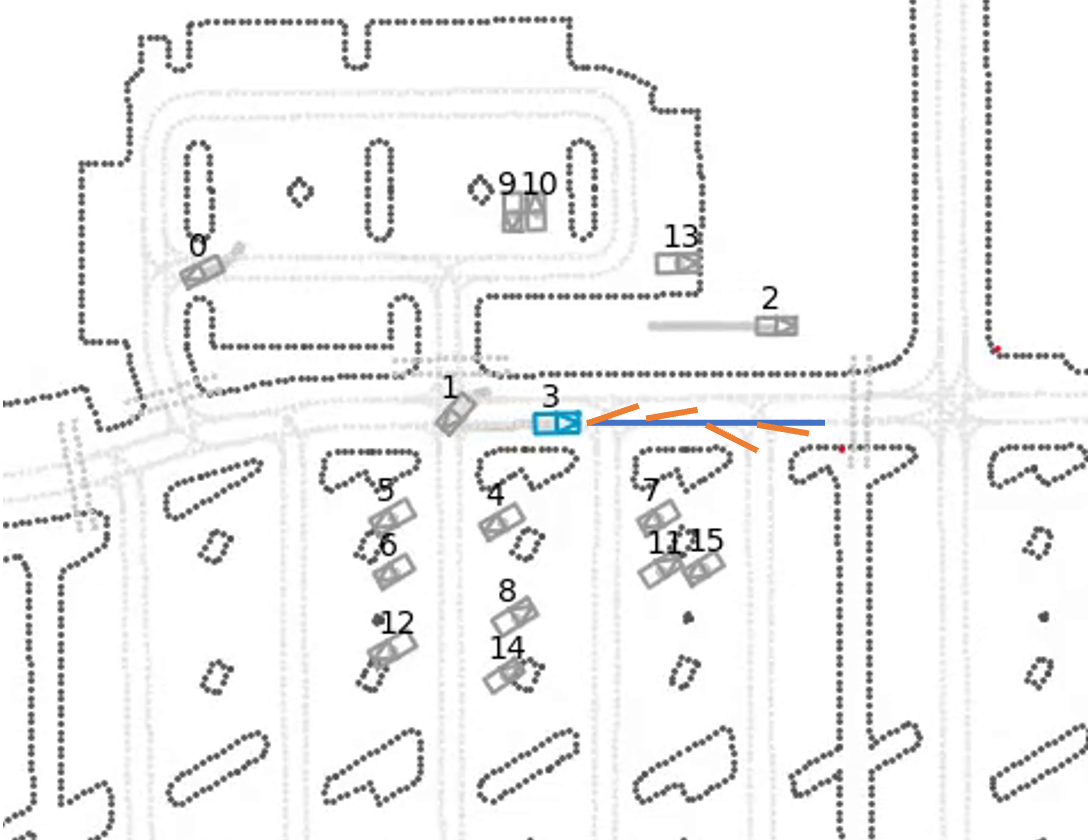}
}
\hfill
\subfloat[T Junction\label{fig:4-3}]{%
\includegraphics[width=0.29\linewidth]{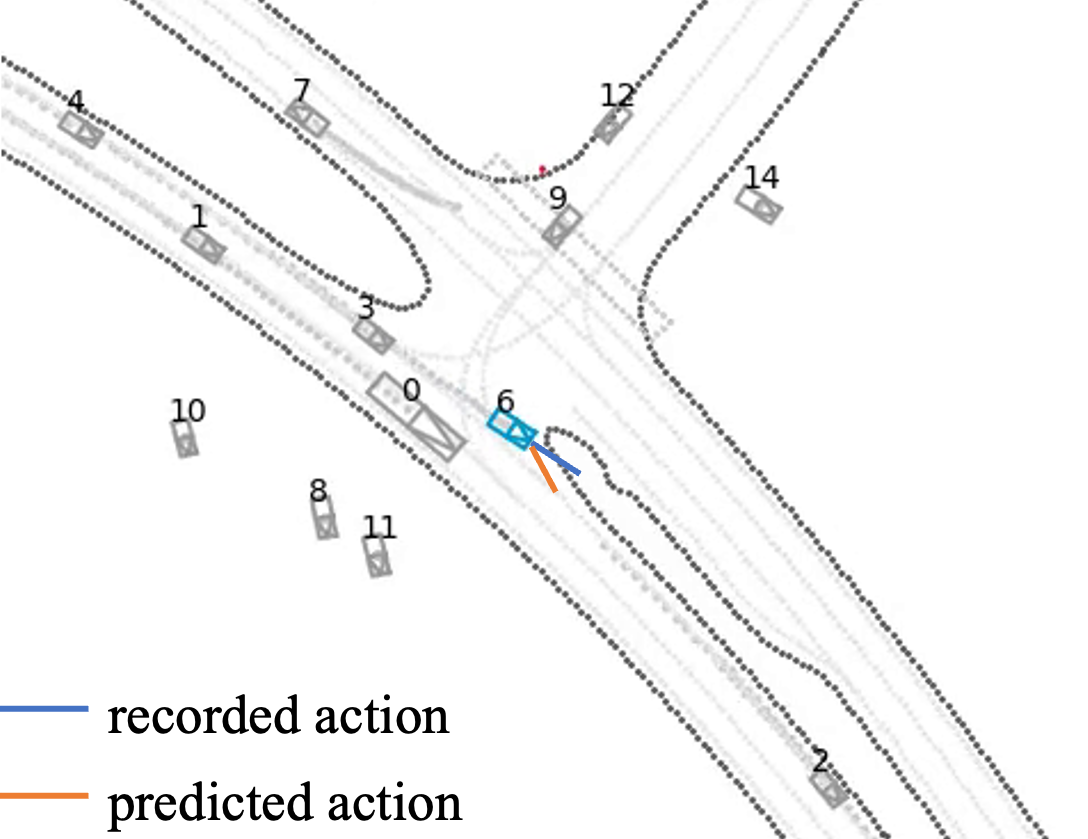}
}
\vspace{-5pt}
\caption{Illustration of Data Selection for Hybrid Experience Sampling: \ref{fig:4-1} is chosen because of its uncommon expert behavior that needs to slow down while steering to the right to keep lane. \ref{fig:4-2} illustrates a rare parking situation and was picked because it had the most mistakes when looking at the whole series of actions. \ref{fig:4-3} is kept as the suboptimal (collision) action taken in that situation was not reproduced given the corresponding low return-to-go.}
\label{fig:4}
\vspace{-15pt}
\end{figure*}

\vspace{-2pt}
\subsection{Benchmark Comparison}
\vspace{-2pt}

PiDT is evaluated in closed-loop settings against both reactive Intelligent Driving Model (IDM) \cite{Treiber_2000} agents and non-reactive playback agents for comprehensive comparison with various algorithms. 

Table~\ref{table:baseline_agents} shows the closed-loop evaluation results when IDM is adapted as the simulated non-ego agent. PiDT achieves an off-road rate of 2.98\%, collision rate of 1.86\%, kinematic infeasibility of 0.00\%, and ADE of 6.99m. PiDT significantly outperforms them in collision rate and is second in off-road rate against other learning-based approaches. Compared with IL methods, PiDT has natural advantage of achieving no kinematic infeasibility. In addition, despite Wayformer \cite{nayakanti2022wayformer} achieving the lowest ADE and LatentDriver \cite{xiao2025latentdriver} achieving the lowest off-road rate, PiDT achieved 82.5\% and 41.3\% reduction in collision rate respectively, which indicates that PiDT learned a safer general driving policy rather than simple imitation. Compared to other RL methods, PiDT shows a substantial reduction in off-road rate and collision rate. PiDT achieves a 60.6\% reduction in collision rate compared to the latest model EasyChauffeur-PPO \cite{xiao2024easychauffeur}, lowering it to 1.86\%. Suggesting that our approach is more effective at keeping the vehicle on the road and avoiding accidents. Compared to traditional IRL method (DQfD) that pretrains with IL and fintunes with RL , PiDT still shows significant improvement with 37.8\% reduction in collision rate and 22.2\% reduction in off-road rate. This improvement in safety-critical metrics highlights PiDT's robustness in real-world driving scenarios.

The closed-loop evaluation against the playback agent (Table~\ref{table:open_loopresults}) shows that PiDT has comparable results with the state-of-the-art IRL algorithm BC-SAC. BC again shows solid performance with a low failure rate of 4.35\% and a high route progress ratio of 99.00\%. BC-SAC improves upon SAC, reduces the failure rate to 3.35\% and increases route progress to 95.26\%. Remarkably, our proposed PiDT method achieves an reduced failure rate of 3.17\% while surpassing others with a route progress ratio of 104.63\%, indicating not only effective navigation but also the potential for discovering more efficient routes. 

\subsection{Ablation Study and Analysis}

\vspace{-2pt}

\noindent\textbf{Effect of Online Imitative Reinforcement Learning}. We evaluate the effectiveness of the OIRL (Alg.~\ref{pipeline}) by comparing PiDT with Wayformer, BC and DQN from Table~\ref{table:baseline_agents}. The reward shaping in Alg.~\ref{pipeline} is crucial for generating the action $a$ and return-to-go $g$ for DT. Without action and return-to-go values provided as input, the network defaults to Wayformer as it directly learns to predict trajectory. Without provided return-to-go by Alg.~\ref{pipeline}, the network defaults to BC as both DT and BC share the same feature encoder. PiDT demonstrates significant performance improvements in several key metrics. Specifically, compared to Wayformer, PiDT achieves a 62.2\% reduction in off-road rate and a 82.5\% reduction in collision rate. When compared to BC, PiDT shows a 28\% reduction in off-road rate and a 68.1\% reduction in collision rate, which indicates enhanced safety and stability in navigation. Against DQN, PiDT provides a 30.8\% reduction in off-road rate and a 62.1
\% reduction in collision rate, the comparative results presented underscore the effectiveness of injecting expert prior to RL. 


\noindent\textbf{Effect of Hybrid Experience Sampling for Decision Transformer}.
 Integrating Hybrid Experience Sampling (HES) into the DT substantially improves performance over the baseline PiDT model. Specifically, the introduction of HES reduces off-road incidents and collision rates by 8.1\% and 9.0\%, respectively, as detailed in Table~\ref{table:ablation}. Furthermore, the enhanced model achieves comparable results using only 60\% of the scenarios, as evidenced by the learning curve comparison in Fig.~\ref{fig:8}. Extensive experiments with various initial seeds confirm the reproducibility of these results. HES strategically stores three categories of data (Fig.~\ref{fig:4}): (1) instances where there is a clear discrepancy between the predicted actions and those executed by experts; (2) trajectories where the cumulative action loss is significantly high which is a phenomenon predominantly observed under rare environmental conditions; (3) cases that reflect suboptimal online adaptation, thereby revealing the model's challenges in discerning and correcting inefficient behaviors.

At a higher level, training on large-scale real-world data leads to an uneven distribution of scenarios. The network effectively generalizes on common scenarios, characterized by low action loss, yet struggles with rare scenarios (illustrated in Fig.~\ref{fig:4-1} and \ref{fig:4-2}), where higher action loss is prevalent. As a result, the replay buffer tends to store more of these challenging instances. To mitigate this imbalance, \hyperref[criterion2]{Criterion 2} is applied, using the accumulated action loss to prioritize rare scenarios. Fig.~\ref{fig:4-3} demonstrates a policy rollout during online adaptation, where suboptimal actions lead to changes in $g_t$. Recording the corresponding $g_t$ for each action is vital: expert actions earn positive rewards, while suboptimal ones incur negative rewards (for example, an ego vehicle drifting out of its lane is associated with a $g_t$ decrease of -2). This detailed tracking of state-action-return sequences equips the neural network to better recognize and correct detrimental behaviors during future encounters with similar scenarios.

\noindent\textbf{Effect of Port-Hamiltonian Neural Network}.
Incorporating the Port-Hamiltonian Neural Network (PHNN) produces the best overall performance among all configurations. Relative to the strongest non-PHNN baseline (PiDT + OIRL + HES), the off-road rate declines from \(3.52\%\) to \(2.98\%\) (about 15\% relative reduction), while the collision rate falls from \(2.69\%\) to \(1.86\%\) (about 31\% reduction). Average Displacement Error (ADE) also decreases from \(7.14\text{ m}\) to \(6.99\text{ m}\). These gains, particularly in safety-critical metrics, indicate that embedding energy-consistent dynamics via PHNN enhances lane adherence and collision avoidance, yielding a more physically plausible and reliable motion-planning policy.

\section{CONCLUSIONS}

We present PiDT, a novel two-stage sequence modeling-based reinforcement learning approach for closed-loop driving and complex real-world driving scenarios. Our online imitative Decision Transformer pipeline sequentially models demonstration and exploration data, jointly optimizes without conflicting objectives. It handles diverse distributions in large-scale driving datasets, ensuring broad applicability and robustness. By implementing a second stage port-hamiltonian neural network, we improve the model's ability to predict over longer horizons and extend its prediction span across broader physics contexts. Furthermore, the incorporation of prioritized experience replay within our framework enhances the sample efficiency of training, allowing for more effective learning from large-scale datasets.

Since the Waymo motion dataset provides only 9-second sequences for each driving segment, PiDT is configured for mid-distance goal-directed tasks that utilize information of 10 historical states and navigate from point $A$ to $B$ within 80 steps. Additional input is necessary for consecutive goal points if we expand it in long-distance driving. The algorithm is also limited memory of past states. These constrains PiDT's long-term reasoning and planning capabilities. We aim to address these issues in subsequent works.

\addtolength{\textheight}{-0cm}   





\bibliographystyle{ieeetr}
\bibliography{references}

\end{document}